# Agentic publications: redesigning scientific publishing in the age of thinking large language models

**Roberto Pugliese and George Kourousias**
*Elettra Sincrotrone Trieste SCpA, Trieste, Italy,*
**and**
**Francesco Venier and Grazia Garlatti Costa**
*University of Trieste, Trieste, Italy*

## Abstract

**Purpose:** This paper introduces the concept of "Agentic Publication," a novel LLM-driven framework designed to complement traditional scientific publishing by transforming papers into interactive knowledge systems that address challenges created by exponential growth in scientific literature.
**Design/methodology/approach:** Our architecture integrates structured data (knowledge graphs, metadata) with unstructured content (text, multimedia) through retrieval-augmented generation and multi-agent verification. The system provides interfaces for humans and artificial agents, offering narrative explanations alongside machine-readable outputs. Implementation leverages vector databases for semantic search, knowledge graphs for structured reasoning, and collaborative verification agents.
**Findings:** Our proof-of-concept demonstration showcases multilingual interaction, API accessibility, continuous knowledge flow, and structured knowledge representation. The framework enables dynamic updating of knowledge, synthesis of new findings, and customizable detail levels.
**Originality:** The Agentic Publication represents a transformative approach to scientific communication by creating responsive knowledge synthesis systems while maintaining scientific rigor. Integrating multi-agent verification with traditional publishing pathways creates a more efficient, accessible, and collaborative research ecosystem, particularly valuable in interdisciplinary fields.
**Practical implications:** The system is a powerful companion for researchers navigating complex knowledge landscapes, offering tailored information access across disciplines while addressing ethical considerations through automated validation, expert oversight, and transparent governance.

## Highlights

- The Agentic Publication transforms static scientific papers into dynamic, interactive knowledge systems powered by large language models
- The architecture combines structured and unstructured data with retrieval-augmented generation and multi-agent verification processes
- The framework provides distinct interfaces for humans and artificial agents, enabling both narrative explanations and machine-readable outputs
- Ethical considerations are addressed through automated validation, expert oversight, and transparent governance principles
- A proof-of-concept demonstrates a practical implementation while preserving compatibility with traditional scientific publishing workflows

## 1. Introduction

Modern scientific publishing, centered on peer-reviewed journal articles, has driven knowledge dissemination for centuries. However, there is growing recognition that the traditional model is strained and increasingly inadequate (Hughes and Van Heerden, 2024). Publication bias and editorial priorities filter what research enters the public domain, while the fundamental format – static documents written for humans to read – has remained unchanged since the 17th century, even as the volume of literature grows exponentially and becomes impossible for any individual to fully process (Hu et al., 2024; Bucur et al., 2022).

This crisis extends beyond format limitations to encompass fundamental systemic challenges in knowledge validation and dissemination.

While the traditional peer-reviewed publishing system has provided essential quality control and institutional credibility for centuries, serving as the backbone of scientific knowledge validation (Kelly et al., 2014), it faces mounting challenges that compromise its effectiveness in contemporary research environments. The exponential growth of scientific literature—with nearly 2 million articles published annually and annual growth rates exceeding 4% (Bornmann et al., 2021)—has created systematic bottlenecks that strain peer review capacity and make comprehensive knowledge synthesis impossible for individual researchers (Tennant & Ross-Hellauer, 2020). This unprecedented growth has led to an already "overstretched" peer review system (Seghier, 2025), creating a cycle where researchers face mounting publication demands while simultaneously experiencing "reviewer fatigue" from increased review burdens (Drozdz & Ladomery, 2024).

The challenge is particularly acute in fast-moving fields like artificial intelligence, where the pace of industry research increasingly outstrips academic publishing's ability to keep up, risking academia's marginalization as a second-tier player in cutting-edge development (Kozak, 2025).

Contemporary challenges are further compounded by the rise of artificial intelligence in academic writing. A growing number of authors now use generative AI to expedite the writing process (Van Noorden & Perkel, 2023), contributing to increased submission volumes that exacerbate existing system pressures. More concerning is the alarming increase in fraudulent manuscripts produced by "paper mills" exploiting AI tools (Van Noorden, 2023), which places additional burden on an already strained verification system. Meanwhile, the availability of qualified reviewers continues to decline, with researchers increasingly declining review invitations due to time constraints and competing professional demands (Willis, 2016), raising questions about the long-term sustainability of traditional peer review in some fields (Künzli et al., 2022).

Beyond scalability issues, the traditional system exhibits documented structural limitations including access barriers through commercial paywalls that create geographic and economic inequalities (Weingart, 2025), publication delays spanning months or years that impede rapid knowledge dissemination, and systematic biases in peer review that can disadvantage research based on author characteristics rather than scientific merit (Lee et al., 2013; Manchikanti et al., 2015). Research has identified multiple bias categories affecting manuscript evaluation, including prestige bias favoring authors from renowned institutions, geographic bias disadvantaging non-Anglophone researchers, and publication bias preferring statistically significant results over null findings (Haffar et al., 2019). Additionally, traditional peer review proves limited in detecting methodological errors and ensuring reproducibility, as evidenced by rising retraction rates and the ongoing reproducibility crisis (Emile et al., 2022). Studies show that reviewers identify only a fraction of deliberately introduced flaws in manuscripts (Drozdz & Ladomery, 2024).

These converging pressures—exponential growth in submissions, declining reviewer availability, AI-driven increases in both legitimate and fraudulent manuscripts—suggest that while artificial intelligence may exacerbate some traditional publishing challenges, it may also offer part of the solution for creating more efficient and scalable knowledge validation systems (Seghier, 2025; Kozak, 2025). The urgency of adaptation is underscored by the risk that scholarly publishing may be sidelined entirely unless it accelerates its processes while preserving quality control (Kozak, 2025). These documented challenges do not diminish the fundamental value of rigorous quality control mechanisms that traditional publishing provides. Rather, they highlight opportunities for complementary approaches that preserve essential peer oversight while addressing structural limitations through technological innovation—a need that becomes particularly acute in interdisciplinary fields where researchers must synthesize knowledge across multiple domains and methodological traditions.

Meanwhile, advances in artificial intelligence and particularly Large Language Models (LLMs) offer a tantalizing opportunity to reimagine how scientific knowledge is shared.

Since the release of ChatGPT in late 2022, awareness and use of LLMs in academia have surged from a niche curiosity to mainstream: a recent survey found that 80% of scientists had used AI chatbots like ChatGPT in their work (Jen and Salam, 2024; Daykan and O'Reilly, 2023; Hughes and Van Heerden, 2024). Initial debates focused on using LLMs as writing assistants to polish or generate papers, raising concerns about plagiarism and accuracy (Lin, 2023; Hughes and Van Heerden, 2024). However, researchers now look beyond AI-assisted writing toward more transformative possibilities (Ahaley *et al.,* 2023; Jen and Salam, 2024; Daykan and O'Reilly, 2023). Some authors claim that next-generation LLMs (guided by human experts) could radically disrupt the entire scientific inquiry and knowledge-sharing system, effectively replacing conventional journals and papers (Ahaley *et al.,* 2023; Hughes and Van Heerden, 2024; Kozak, 2025). Instead of engaging solely with static documents, researchers and interested stakeholders could interact with dynamic knowledge repositories that integrate published findings and provide contextual information responsively. For social scientists, this might include up-to-date economic indicators, policy analyses, and management research syntheses, enabling more informed decision-making through timely access to relevant knowledge. Similarly, natural scientists could benefit from real-time integration of experimental results across laboratories, creating opportunities for accelerated discovery through improved knowledge coordination. In this vision, the traditional paper could soon appear outdated (Hughes and Van Heerden, 2024).

Building upon these opportunities and addressing the gaps in scientific dissemination, this manuscript proposes an LLM-powered architecture as a potential complement to the traditional scientific paper, which we term an Agentic Publication (AP). An AP is a scientific artifact enhanced with embedded artificial agents powered by large language models, which dynamically integrate, verify, and disseminate knowledge. This approach has implications across scientific disciplines, including both natural and social sciences, where the need for dynamic knowledge synthesis is particularly acute.

Unlike traditional papers, APs are interactive, continuously updatable, and accessible to both human users and other machines. Our goal is outlining a system – composed of large language models, tools and intelligent agents – that can support scientists to create, store, process, and interactively disseminate scientific knowledge on demand. We outline the structure of this system, including its core components (knowledge representation, query interface, updating and verification modules), a potential technology stack, and implementation strategies. We also discuss how the system can present information differently to human users versus artificial agents and examine the associated ethical implications. By detailing this architecture and its feasibility, we aim to illustrate a path toward a more accessible, up-to-date, and interactive model of scientific communication (Hughes and Van Heerden, 2024) – open and responsive, benefiting researchers and society in ways the static journal article alone cannot.

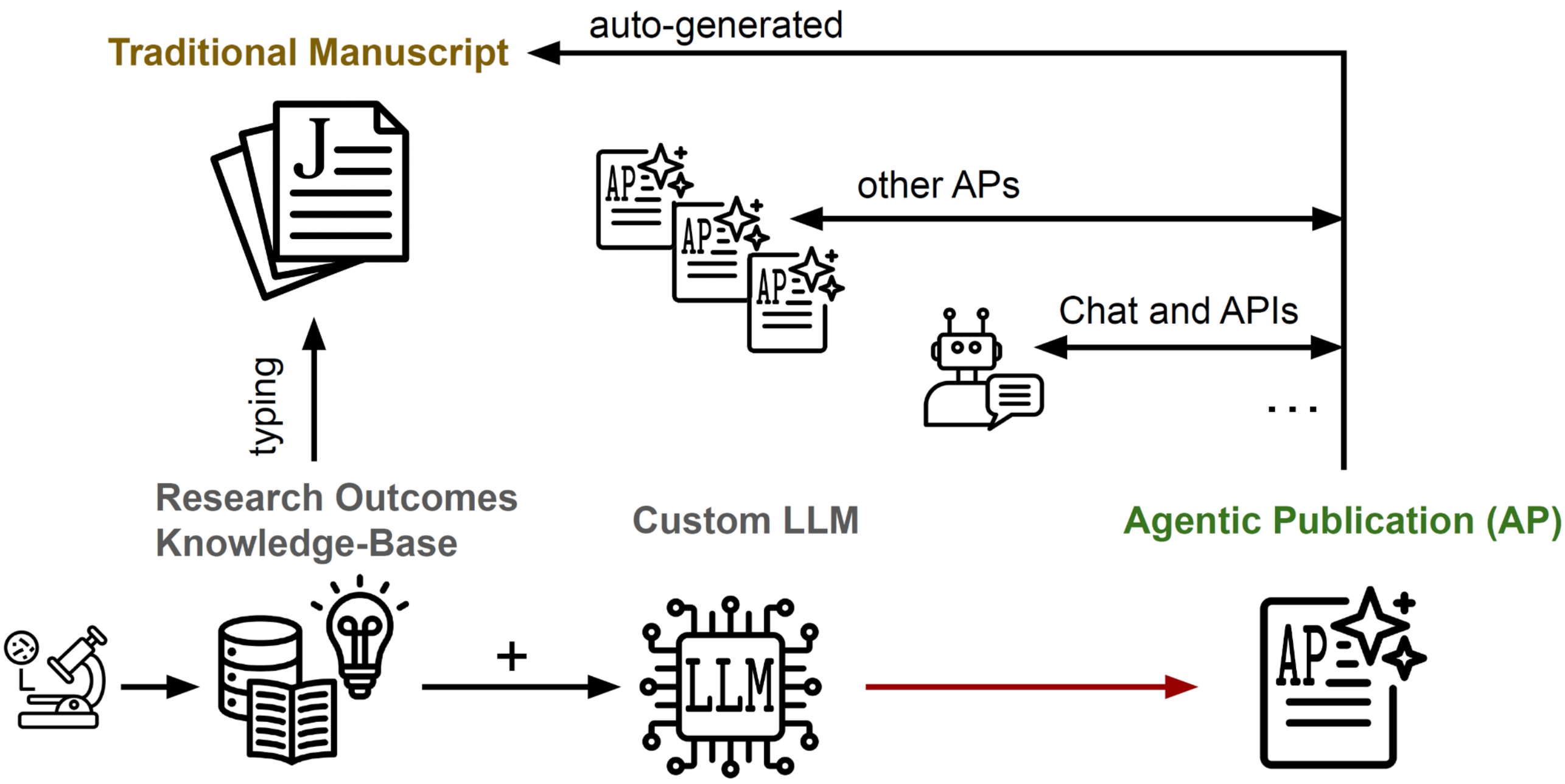

*Figure 1: Simplified overview of how research outcomes yield in a traditional manuscript and an Agentic Publication (AP). The knowledge-base including structured and unstructured content (text) is coupled with a suitable AP LLM considering the AP properties described in this manuscript (red arrow) resulting in an AP. APs can be used for basic "chat with your paper" but also provide APIs for computational access (inc. to the data). APs can interact with other APs but also auto-generate print-ready versions of traditional manuscripts. These auto-generated manuscripts are outputs of the AP for human consumption and accessibility, but the AP itself remains the primary citable entity.*

## 2. Proposed System Architecture

### *2.1 System Architecture Overview*

The proposed system is centered on a large language model continually enriched with scientific findings. We term this concept "Agentic Publication" as it includes AI agent features and supports automated research processes. Figure 1 shows the system's conceptual overview, encompassing new research information flow and knowledge base querying.

The architecture (see Figure 2 and Figure 3) comprises four interacting components: (1) a *Knowledge Representation Layer* storing scientific information in human-readable and machine-interpretable forms; (2) an *Interactive Query Interface* (API, chat, or voice-based) for user questions and responses; (3) *Dynamic Updating Mechanisms* that ingest new findings based on query analysis; and (4) *Verification and Governance Processes* ensuring knowledge base quality and integrity. These components are supported by advanced LLMs, agents, databases, and integration APIs.

While Figure 1 illustrates that APs can auto-generate print-ready manuscripts, these should be understood as outputs derived from the AP rather than as the AP itself. An AP consists of the curated knowledge base, author analyses, and embedded agentic mechanisms with version tracking. From this foundation, multiple manuscript variants can be produced according to author-defined requirements such as style, language, or length. To ensure consistency, authors may pre-approve the generation prompt or provide pre-generated, curated versions stored within the AP. In this way, the AP remains an inherently dynamic and interactive system, while also enabling the export of fixed-form documents for human consumption. Auto-generated manuscripts should not be regarded as peer-reviewed, citable works in their own right; rather, the AP is the primary referential entity, with generated documents serving as associated resources that support accessibility and dissemination.

### *2.2 Knowledge Representation Layer*

Scientific knowledge representation across domains requires formats suitable for both LLM and algorithm consumption, accommodating varied methodologies of different fields.

We propose a hybrid knowledge store with complementary components working as an integrated system. At its foundation lies unstructured content comprising full research texts, including methodology descriptions, observations, and results, along with associated artifacts such as figures, tables, and datasets. The content is indexed in vector databases for semantic search, enabling LLM retrieval of relevant passages. Semantic indexing captures conceptual relationships despite terminology differences (Hughes and Van Heerden, 2024). Each knowledge entry carries important metadata including authors, publication date, and domain keywords to provide essential context.

Working in parallel with this unstructured foundation, structured knowledge components store key facts and relationships in knowledge graphs (Wang and Shi, 2025) or relational databases. This layer employs ontologies to define entities ranging from chemical compounds to business models, enabling machine-interpretable encoding and consistency checks across studies examining similar relationships. Other scientific data formats like (HDF) can be supported too, in case via integration tools.

Building on semantic publishing work (Bucur *et al.,* 2022), our system uses LLMs to bridge structured and unstructured realms, interpreting data while populating knowledge graphs.

Multi-modal data support addresses scientific knowledge including datasets, code, images, and audio/video recordings. Multi-modal LLMs can accept and generate multiple data types. The architecture incorporates data repositories linking supplementary materials with textual descriptions, enabling detailed responses and verification against source data. Vision-language models may interpret figures and experimental diagrams, enhancing knowledge integration across disciplines (Dias *et al.,* 2023).

Provenance tracking documents evolution with transparency, preserving original LLMs, reviewer comments, revision history, and updates. This strengthens scientific accountability and reproducibility.

In addition to datasets and results, APs are designed to explicitly embed the authors' professional reasoning: interpretations, critiques, links to other fields, and reflections on limitations. This "authorial reasoning layer" ensures that the AP is not merely *data + LLM*, but rather *data + LLM + authorial analysis*. Authors can also provide reasoning profiles or bias prompts that guide the system's responses in transparent and reproducible ways. For example, an author might specify: *"all LLM reasoning should be compatible with Epicurean views," "all responses should remain consistent with the author's previous work,"* or *"ignore string-theory approaches when formulating answers."* These profiles are optional, versioned, and signed (e.g., ORCID-linked), so that readers and downstream systems can explicitly distinguish between evidence-based outputs and author-guided perspectives. In this way, opinions and viewpoints remain first-class scholarly contributions in APs, complementing the underlying evidence and making reasoning both auditable and reusable.

The architecture's design incorporates upgradability as a fundamental principle, giving an Agentic Publications inherent potential for continual improvement and upgrading. Upgradability allows seamless integration of new LLMs, enhancing analytical capabilities and engagement beyond original constraints, ensuring long-term relevance.

By combining these, the knowledge representation layer serves as system memory and truth source. It provides redundancy (text for flexibility; structured data for precision) minimizing errors. Scalable infrastructure will support millions of entries using existing "big data" and datalake technologies (Huang, 2021).

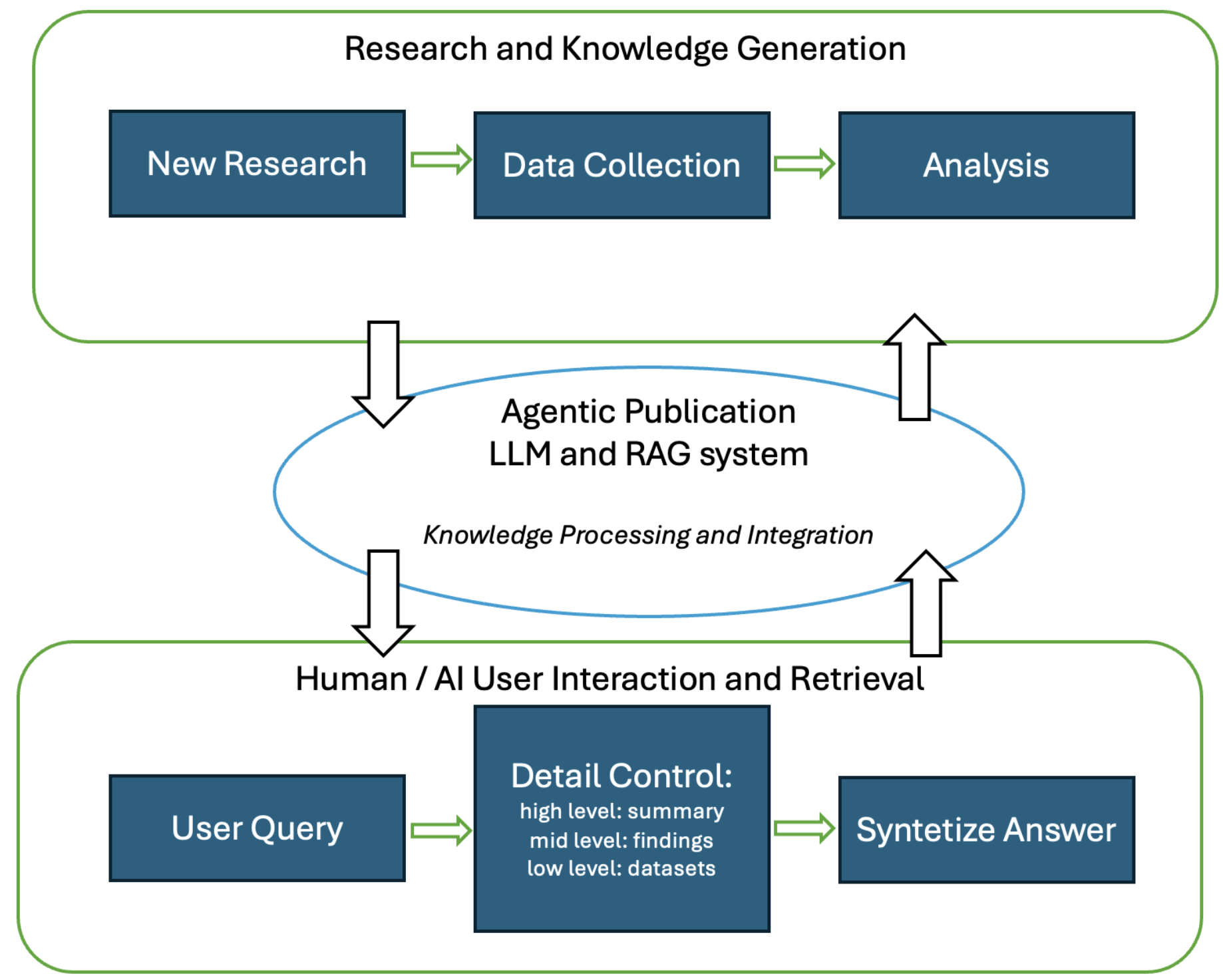

*Figure 2: Conceptual illustration of the proposed Agentic Publication (inspired by the "PLOS-LLM" model by Hughes and Van Heerden, 2024). New research and knowledge generation steps feed data into the LLM-centric system, while user queries retrieve synthesized answers. The system allows users to zoom in and out on the level of detail – from high-level summaries (headlines/abstracts) to granular data (complete datasets). An interactive loop based on interaction log analysis helps keep knowledge updated and easily accessible.*

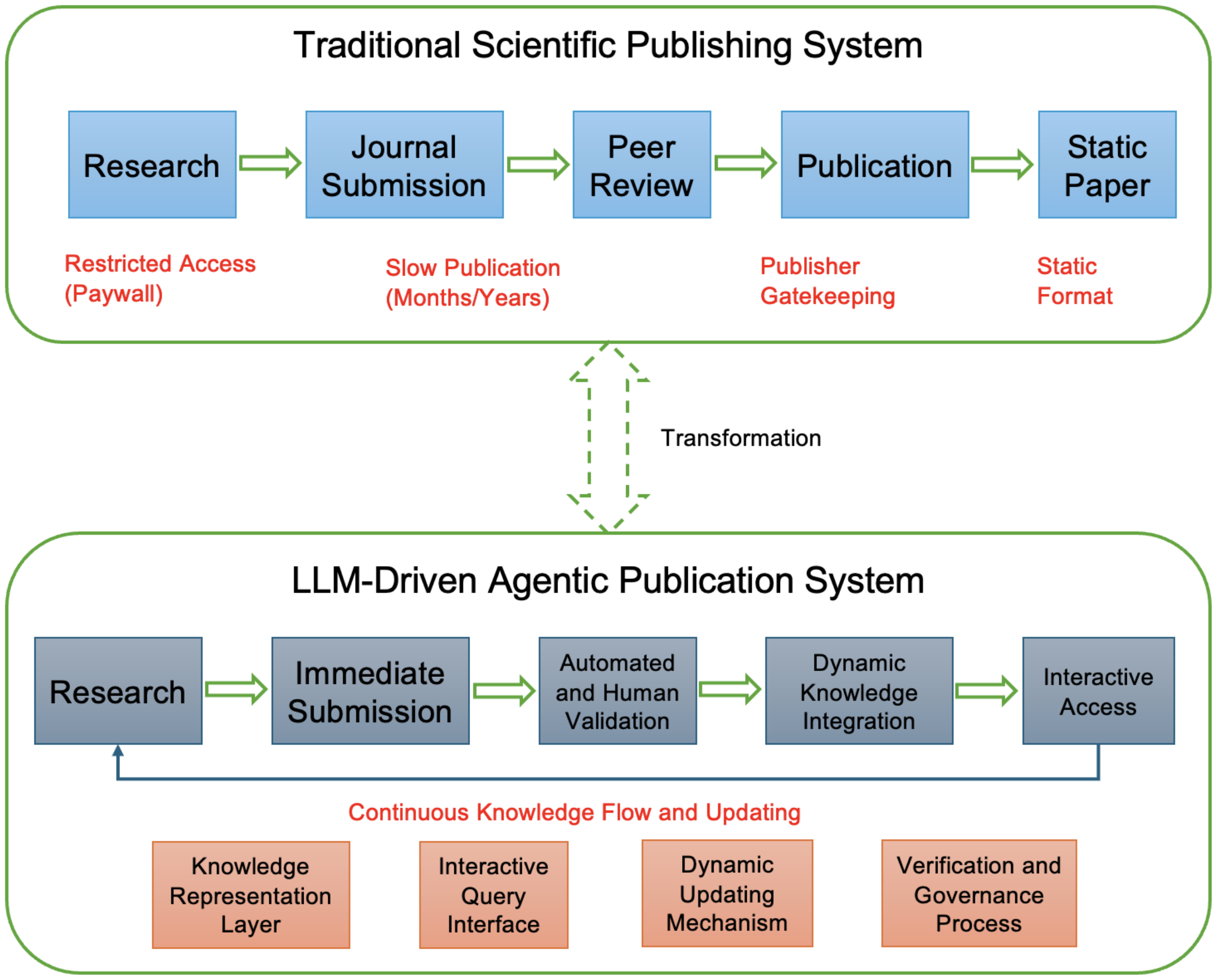

*Figure 3: Scientific Knowledge Workflow: Traditional versus LLM-Driven Agentic Publication System.*

### *2.3 Interactive Query Interface*

Users interact with the knowledge system through an intuitive conversational interface, enhancing static article utility (Li and Zhang, 2023). Users pose questions in natural language—*"What are the latest findings on X?"* or *"Show me data supporting claim Q"*—and receive tailored answers synthesized from relevant knowledge in the same language of the request.

The interface design incorporates several essential features that enhance scientific knowledge accessibility. Conversational Q&A stands at the core of this system, supporting iterative dialogue with follow-up questions and clarification requests. The LLM contextual understanding enables natural conversation flow, offering greater flexibility than keyword searches by integrating information across sources and tailoring explanations to specific questions.

Another crucial aspect is adjustable detail functionality, often called "zoom", which provides varying information levels for different users. Users can access succinct summaries, abstract-style overviews, detailed discussions with evidence, or raw experimental data (Hughes and Van Heerden, 2024). For instance, policymakers might need consensus conclusions, while experts require underlying data inspection. This dynamic adjustment replaces static paper text with responsive presentation tailored to user needs.

The system further enhances knowledge delivery through multi-modal responses that leverage diverse knowledge base data. When asked about trends, the system generates graphs from datasets, analyzes images, and integrates tables, images, or audio with text responses, making knowledge consumption more engaging (Cardon, 2023).

User-friendly input mechanisms further lower usage barriers including voice interaction through speech-to-text, enabling hands-free operation valuable for practitioners in field settings. Web and mobile platform availability democratizes scientific knowledge access (Li and Zhang, 2023).

Finally, context and personalization features enhance the user experience by maintaining user profiles, adjusting explanations based on background—simpler language for students, technical detail for experts—while remembering previous interactions to avoid redundancy.

Behind this interface, the LLMs and associated agents do the heavy lifting, interpreting the question, fetching relevant knowledge, and composing an answer. Answers include references or links to supporting knowledge base data. For instance, the interface might show in-line citations that users can click to inspect the original study or data source that backs each statement. Combining natural language with ease with scholarly rigor (sources, evidence on demand), the interactive interface can make consuming scientific knowledge across disciplines convenient and trustworthy.

### *2.4 Agents in an Agentic Publication*

An AP can be also described as an Agent in an agents ecosystem. In the context of Large Language Models (LLMs), an agent refers to an AI system that leverages an LLM as its "brain" to autonomously or semi-autonomously perform complex tasks, make decisions, and interact with users or other systems. Unlike a standalone LLM—which simply generates text responses to inputs—an LLM agent can plan actions, use external tools, and maintain memory across interactions, enabling it to handle multi-step workflows and dynamic environments.

Key characteristics of LLM agents include autonomy, planning, tool use, memory, reasoning and persona/profiling. It is anyhow reasonable to think about an agent ecosystem, where scientists can create new AP agents by uploading their specific knowledge and discoveries. The ecosystem also includes non-AP agents that are dedicated to specific bookkeeping tasks to support the functioning and integrity of the whole AP ecosystem. In the rest of the paper when appropriate, we will specify if the presented features will be implemented by the AP agent or by these non AP ecosystem agents.

As it is now evident, APs can include multiple agents, each fulfilling distinct roles. For example, a primary authoring agent may synthesize content, a verification agent can cross-reference claims, and a reviewer agent may simulate critical peer review. This modular design fosters collaborative intelligence and robust internal checks, improving trust and scalability.

***2.5 Dynamic Updating Mechanisms***

A cornerstone of our proposed system is that it remains up-to-date, reflecting the evolving nature of scientific knowledge across all fields (Klami *et al.,* 2024). Rather than the multi-month or multi-year cycles of traditional publishing and literature reviews, this architecture allows scientific knowledge to be continuously refreshed in real time.

The foundation of our approach relies on continuous knowledge ingestion, where researchers submit findings to the system as soon as they are available, rather than waiting for complete, polished papers. Once a study's methodology is approved and experiments are conducted, the methods and initial results can be uploaded to the AP ecosystem as a new agent (Hughes and Van Heerden, 2024). This approach can even precede formal analysis – early hypotheses, intermediate results, or registered trial protocols can be fed in, so the model knows "something is in progress." When final results are obtained, authors upload those along with datasets and analyses. These parallels open notebook science principles, where the knowledge base is incrementally built as research progresses, coupled to an intelligent model.

Supporting this continuous flow are automated pipelines and APIs that facilitate ingestion by accepting various input formats. Researchers might fill structured templates for their studies or upload preprints for system parsing. Integration with existing databases is crucial – connecting to arXiv or PubMed to automatically pull new content daily. Journal platforms could evolve to interface directly with the AP ecosystem, pushing content upon publication. Web crawlers can scan repositories for relevant updates, with all inputs time-stamped and labeled for version control.

To ensure reliability, each submission triggers a validation routine before full integration. This involves automated checks executed by bookkeeping agents (plagiarism detection, consistency verification, statistical review) and routing to human experts when needed. Only after passing this validation gate is knowledge considered verified in the system, ensuring the model's knowledge base remains reliable and does not accumulate unchecked errors.

The core architecture incorporates model updating processes occurring in two ways. A retrieval-augmented generation approach immediately incorporates new facts by referencing the latest information from the AP knowledge repository at query time. However, periodic fine-tuning of the LLM improves fluency and reasoning on new data – the model might be updated nightly or weekly on new content. This dual approach balances currency and performance while mitigating context window limitations (Hu *et al.,* 2024).

A powerful feature is the AP system's capacity for automated synthesis and updates, where aggregated knowledge updates automatically with each new piece of evidence. Meta-analyses could re-run as new data arrives, updating consensus answers and recomputing summary statistics. This means reviews never go out of date – a stark contrast to static articles obsolete upon publication (Hughes and Van Heerden, 2024). The system maintains uncertainty estimates that update as evidence accumulates.

Finally, the architecture addresses scalability requirements as the knowledge base grows. It might be partitioned by domain or distributed globally, with specialized AP sub systems instances for different fields sharing backbones while receiving domain-specific fine-tuning. These strategies ensure responsiveness even as thousands of studies integrate.

In summary, dynamic updating mechanisms transform scientific knowledge dissemination from a slow, discontinuous process into a live, continuous, and responsive cycle. New knowledge is ingested promptly, validated, and made queryable, with the model's outputs evolving as knowledge evolves (Klami *et al.,* 2024). This closes the gap between discovery and dissemination, accelerating insight spread through scientific communities.

***2.6 Verification and Quality Control Processes***

Any system disseminating scientific knowledge must uphold accuracy, credibility, and trustworthiness standards. While traditional peer review serves as a quality gatekeeper, it has well-documented flaws and delays. In our LLM-agent architecture, verification is equally crucial, given AI's propensity to confidently generate incorrect information if unchecked. We propose a multi-layered verification process combining human expertise and automated agents.

Central to our verification framework is human expert oversight through "AI-assisted peer review." When submitting a new AP, the system identifies relevant experts using its knowledge graph, inviting them to evaluate submissions (Hughes and Van Heerden, 2024). Experts interact with submissions via natural language queries: "How was sample size chosen? Are there similar studies to compare?" The AI-assisted review surfaces issues quickly by comparing new results to related literature and highlighting conflicts for reviewer attention. Reviewers provide feedback and scores rating rigor, novelty, and validity. Only studies meeting credibility thresholds are fully integrated, with others tagged with caution or lower confidence scores.

This process is transparent: reviews and scores are logged and attached to knowledge entries for public viewing, unlike traditional hidden peer review. Over time, reputation systems could develop where contributors and reviewers have track records, incentivizing quality contributions. Human oversight remains vital to catch nuances AI might miss (Hughes and Van Heerden, 2024).

Automated validation agents enhance verification through specialized AI performing specific checks. Evidence Validator LLMs cross-check factual claims against existing knowledge, while statistical checkers re-calculate results and flag analysis errors like p-hacking. Provenance tracers verify citations and guard against plagiarism. Multiple agents critique submissions in seconds, functioning as exhaustive reviewers that check everything from mathematics to consistency with known physics (Hughes and Van Heerden, 2024).

The system implements continuous error monitoring at query time through secondary processes that verify responses before display. When the LLM produces summaries, retrieval steps pull original sources to ensure accuracy. If stating "Study X found Y," the AP system double-checks that Study X indeed reported Y, mitigating hallucinations and building user trust.

A critical aspect of our verification approach involves the calibration of uncertainty as part of quality control, which attaches confidence scores to answers based on source agreement, data quality, and conflicting evidence. When questions lack sufficient data, this is explicitly stated rather than guessed, paralleling how human authors acknowledge inconclusive evidence (Hughes and Van Heerden, 2024).

Complementing these verification mechanisms are safety and bias checks that scan for problematic content. When data overwhelmingly represents specific regions or demographics, potential skew is noted.

Therefore, maintaining quality in an LLM-driven knowledge repository requires reimagining peer review as integral and adaptable to different scientific disciplines. By combining human wisdom with artificial vigilance, the system ensures reliable knowledge dissemination. While LLMs may increasingly catch errors autonomously (Binz *et al.,* 2025), human judgment remains crucial for nuanced evaluations (Klami *et al.,* 2024; Hughes and Van Heerden, 2024).

### *2.7 Technology Stack and Integration*

Building the above system is a significant engineering challenge, but many building blocks exist today. Here, we suggest a plausible technology stack and frameworks to implement the proposed architecture, leveraging state-of-the-art tools.

At the center of our implementation architecture stands the Large Language Model, forming the core intelligence of the system. This could be based on transformer architectures with capabilities on par with GPT-4 or similar models like Gemini 2.5, Claude 3.7, given their demonstrated performance on diverse knowledge tasks (Freire *et al.,* 2024). Organizations might choose an open-source LLM (such as LLaMA 3, DeepSeek, …) and fine-tune it on scientific text to create a domain-optimized model. Techniques like LoRA (Low-Rank Adaptation) can efficiently update the model on new data without full retraining, allowing continuous learning. While closed APIs like OpenAI's GPT-4 could rely on retrieval augmentation exclusively, open models offer more flexibility for direct tuning and on-premises deployment necessary for data privacy. A hierarchy of models might be employed: a large general model for understanding queries and generating answers, with smaller specialized models for specific domains like code execution or image analysis.

Supporting the LLM, a robust retrieval and database layer combines vector databases and traditional databases to implement semantic search and handle diverse data storage needs. Tools like FAISS, Pinecone, or Weaviate can store high-dimensional text embeddings, allowing similarity search to fetch relevant contexts in milliseconds. For structured metadata and knowledge graphs, graph databases like Neo4j could store relationships and enable complex queries. Traditional SQL databases can handle tabular experimental results. The system includes an indexing pipeline: when new data arrives, embeddings are generated and inserted into vector databases, structured facts are extracted for graph databases, and raw files are stored in distributed systems with links.

To coordinate various processes, orchestration and agent frameworks like LangChain, Haystack or CrewAI provide abstractions for chaining LLM calls with retrieval and other actions. For instance, a chain handles queries: user question → LLM parses question → retrieval tool fetches relevant chunks → LLM composes answers. If answers require visualization or analysis, the system could use agents with Python execution tools to load datasets and produce charts. The multi-agent scenario can be coordinated via controller scripts or shared memory systems for agent communication (Hu *et al.,* 2024).

The user experience is delivered through a front-end interface implemented as web or mobile applications. Modern frameworks (React, Vue) could build chat UIs with features like citation highlighting, result filtering, and data visualization outputs. For voice interaction, integration with speech-to-text services and text-to-speech for replies can be added. The front-end communicates with backends via REST or WebSocket APIs, with an emphasis on UX design to present complex scientific answers in an understandable manner.

For external connectivity and system expansion, APIs expose the system's knowledge for external tools and agents to consume. Other AI systems might query this knowledge base via APIs supporting natural language queries and returning structured results. Integration with ID systems like DOIs for datasets and ORCID for authors helps maintain links to the broader research ecosystem. Additionally, connecting to existing scientific infrastructure (clinical trial registries and data repositories) via their APIs can enrich the system's content and reliability.

Behind the scenes, a continuous deployment and scaling approach based on cloud microservices architecture hosts these components. LLMs run on GPU servers while databases operate on scalable clusters. Kubernetes can manage services and scale them based on load – during heavy query periods, more instances spin up, while ingestion services scale when batches of new studies arrive. Logging and monitoring services are critical, with every answer logged along with sources used to audit model behavior and detect systematic errors.

To ensure ethical use, model alignment and safety incorporate an AI ethics layer that could implement tools to detect toxic or biased content (Koçak, 2024). This includes intermediate filters checking outputs and adversarial testing suites that regularly probe the system with corner cases or misinformation to refine model responses.

Traditional publications historically required minimal resources, limited to basic web hosting. In contrast, Agentic Publications (APs) have substantially higher cloud resource requirements, necessitating sophisticated infrastructures capable of storing and executing LLMs, maintaining extensive metadata, and dynamically responding to user interactions. The required computational power scales with concurrent readers, introducing significant hosting considerations. Frameworks such as the European Open Science Cloud (EOSC) are ideally suited for hosting APs. Specialized EOSC

Agentic Publication Nodes can be established by major research institutions to make necessary hardware resources available to the research community. The Italian synchrotron Elettra Sincrotrone Trieste is currently experimenting with such infrastructure based on its Data Lake ecosystem. The demonstration Agentic Publication accompanying this paper is hosted through Elettra's cloud infrastructure, expected to yield valuable insights into infrastructure costs, sustainability, and scalability.
Therefore, while the overall system is ambitious, it can be constructed by integrating existing technologies: LLMs for language understanding, vector search for knowledge retrieval (Hu *et al.,* 2024), knowledge graphs for structured reasoning, and orchestrators for tool use. Early prototypes might start with simpler versions and progressively add complex features as components mature in reliability.

## 3. From Concept to Prototype: Implementing the Agentic Publication System

We now outline how one would implement a prototype of this LLM-agent-based knowledge dissemination system, highlighting considerations for diverse scientific fields. The process can be broken down into several stages, from data ingestion to user interaction.

### *3.1 Knowledge Ingestion Pipeline:*

The first step involves assembling the knowledge base through integrated processes that collect and structure research content.
The pipeline begins with data acquisition, where we gather scientific documents from repositories (arXiv, PubMed, Web of Science, Scopus) using tools like Semantic Scholar API or CrossRef for DOI content. This possibility may be useful at the beginning to facilitate system adoption.
A parallel submission portal allows researchers to directly upload manuscripts or research details through web forms.
Once data is acquired, parsing and processing transform raw documents into structured content using NLP techniques. Documents are partitioned into sections (abstract, methods, results) and further into manageable chunks. We can employ PDF parsing libraries or LaTeX source when available to extract text and references, followed by cleaning procedures to remove artifacts and convert specialized notation to plain text.
Alongside content extraction, metadata extraction identifies critical contextual information—title, authors, affiliations, keywords, publication venue, date, and references—using specialized libraries, or DOI-based services. This metadata populates structured databases and enables proper citation formatting.
To enable semantic search capabilities, embeddings indexing converts content chunks into machine-understandable representations using pre-trained models or sentence-transformers fine-tuned on scholarly data. These vectors are stored in vector databases with pointers to original sources, enabling semantic similarity search based on meaning rather than keywords.
For more structured knowledge representation, knowledge graph population creates semantic relationships through named entity recognition (NER) to identify entities (diseases, chemicals, instruments) and relation extraction to detect relationships ("X treats Y"). The LLM can extract structured claims using prompts like "List main claims as triples (subject, relation, object)" with results linked to existing ontology terms for consistency.
Finally, multimedia handling addresses non-textual components by storing original files while generating thumbnails for images and extracting captions as indexable text. Datasets in common formats are loaded into databases or have their schemas recorded for future queries.
This pipeline operates continuously to incorporate new submissions and updates, ensuring current knowledge base maintenance. AI Agents can be programmed and used to implement this pipeline.

### *3.2 LLM Integration and Training*

With data in place, a possible next step is integrating the core LLM through interconnected processes that transform a general model into a specialized scientific knowledge system.
The journey begins with initial training to adapt the model to scientific content. If starting from a pre-trained model (like GPT-3/4), one might fine-tune it on curated and diverse domain data, ensuring representation across scientific disciplines. This could involve supervised fine-tuning on question-answer pairs derived from the corpus or instructing it on knowledge usage by providing examples of how to extract information from papers and answer questions. When fine-tuning data is sparse, an alternative is few-shot prompting at runtime, which allows the system to learn from limited examples while maintaining flexibility. This step may become less important as the quality of the LLMs increases.
Once trained, implementing a retrieval-augmented setup creates the crucial connection between the LLM and knowledge base. This mechanism ensures that before answering queries, the system performs vector search on indexed knowledge. For example, using the LangChain framework, create a QA chain: user query → embed query → similarity search top N passages from vector database → feed those passages plus the question into the LLM prompt as context → get answer. This step is crucial for grounding LLM answers in the specific literature and preventing hallucination. The prompt template should instruct the model to quote sources and use retrieved text to justify answers, maintaining scientific rigor.
To extend capabilities beyond simple question answering, tool use configuration enables the LLM to perform specialized functions when needed. This may use predefined prompt formats (like the ReAct framework or OpenAI function calling) where the model outputs special tokens to call tools such as calculators or plot generators. For instance, if the query is numerical ("What's the average result across these studies?"), the model might call a calculator tool with values extracted from sources, then incorporate the result into its answer. This functionality allows the system to perform computations and generate visualizations that enhance response value. New interoperability standards like the Model Context Protocol (MCP) by Anthropic simplify such integrations by allowing models to interface with diverse tools and environments seamlessly (Hou et al., 2025). MCP helps ensure consistent context handling across agent interactions, enhancing performance in multi-agent AP ecosystems.
For complex queries requiring multiple perspectives or verification, agent collaboration establishes communication channels between specialized components. If implementing multiple agents (like a verifier agent), it is important to configure them to communicate with each other. For example, the main answering agent could pass its response to a verifier agent with knowledge base access for comments or corrections, returning the final answer only after verification. This can be orchestrated through simple loops: generate answer → verify → if verifier finds issues, revise answer. Such collaborative architecture mimics the peer review process in science, enhancing reliability. Inter AI agent communication protocols like the Agent2Agent Protocol (A2A) developed by Google and Microsoft, or the Agent Communication Protocol (ACP) developed by IBM Research, may facilitate this step.
Before deployment, rigorous testing evaluates the system's performance and safety. Begin with known questions that can be verified manually, checking if answers are correct and if provided sources match the content. This is an iterative process: if the model makes mistakes (e.g., attributing results to wrong papers), adjust the prompting, improve retrieval by increasing retrieved passages or adding re-ranking steps, or add fine-tuning examples addressing specific error modes. It is important to ensure the model doesn't return sensitive or inappropriate content. Thorough testing builds confidence in the system's reliability before it faces real-world scientific inquiries.

### *3.3 User Interface Development*

Developing the front-end requires thoughtful interface design balancing functionality with accessibility.

At the core of this interface lies the chat interface, which features a simple chat window with message history for context maintenance and follow-up questions. Citation display makes references like "(Smith et al. 2021)" clickable with complete reference details in tooltips. Rich media content—images, graphs, model-generated plots—renders seamlessly within chat flow for enhanced comprehension.
To accommodate varying information needs, the system should provide detail level controls for detail level through intuitive UI elements such as sliders or buttons that adjust between "summary" and "detail" views. These trigger different prompts requesting brief answers or comprehensive responses with methodological information. While users can request detail levels through natural language, visible UI cues enhance discoverability of advanced functionality.
Accessibility considerations lead us to incorporate voice input/output capabilities that expand interaction beyond text through microphone input with speech transcription and text-to-speech conversion for responses. This enhances accessibility and enables hands-free operation in laboratory or field environments.
Beyond human interfaces, the system should implement an agent API that enables programmatic interaction through REST endpoints (e.g., POST /query) accepting JSON payloads with questions and authentication credentials. Responses return structured data including answers and references, allowing external software and AI agents to consume knowledge programmatically. Comprehensive documentation facilitates integration by developers and other systems. Of course, protocols like A2A or ACP or other emerging standards can be adopted instead.

### *3.4 Dynamic Update & Feedback Loop*

Maintaining system updates and improvements requires comprehensive interconnected components. Scheduled updates implement schedulers (cron jobs) triggering ingestion pipelines for new content daily or weekly. Real-time processes handle direct author submissions with automated re-training or index updates following data ingestion. Continuous integration tools run ingestion or other forms of bookkeeping scripts with monitoring systems catching process failures.
User feedback integration allows answer rating and issue flagging (thumbs up/down, problem flags) with mechanisms capturing domain-specific feedback. This provides valuable retraining input—frequently flagged incorrect answers guide investigation and model improvement, potentially revealing misunderstood nuanced questions requiring additional training data.
Ensuring optimal functioning requires robust performance monitoring logs metrics including response time, queries answered, and retrieval success rates. Query distribution tracking provides strategic insights—frequently asked topics may indicate content gaps or needs for dedicated sub-models.
As the system gains adoption, scaling considerations address growing usage through service replication, load balancers for APIs, multiple LLM container instances, and autoscaling policies. Caching common queries improves efficiency while requiring careful design to handle knowledge updates that change answers, potentially using short TTL or caching only unchanged underlying data.
Throughout the implementation, iterative testing and refinement recognize this as a continuously improving system rather than a finished product. Initial user studies guide development by observing scientist interactions, similar to studies testing LLM tools for workers who appreciated quick retrieval while valuing human expertise (Freire *et al.,* 2024). Such feedback informs further development, including explanatory features or verification tightening based on trust concerns.
The result is a working prototype where users receive coherent answers with sources, new papers influence responses, and the system maintains conversational interaction through continuous improvement.

### *3.5 Agentic Publication Demo*

To examine and support implementation's feasibility, a limited demo has been developed and deployed. This demo implements a single AP without any AI bookkeeping agent and is not meant to demonstrate all the features of a complete agentic publication system, since it has already been established that this requires targeted analysis and professional software engineering. This manuscript aims to provide the concepts, architecture and insights on the Agentic Publication concept. That said, the authors coded with RAG principles a limited single Agentic Publication like version with as the knowledge-base (kb) the very same content of this manuscript. The Agentic Publication demo is accessible and has been assigned a DOI (https://doi.org/10.34965/agenticpublication.3567a) (Pugliese, 2025) and its features include a *landing page* that presents the paper metadata and a description of the different features currently available. The current version of the demo has been implemented as a Chatbot based on the VoiceFlow platform and implementing a set of workflows that define the basic behaviour of the Agentic Publication.

Voiceflow is a comprehensive platform for designing, prototyping, and deploying conversational AI applications. The platform enables the creation of voice and chat interfaces through a visual, no-code interface that allows developers and non-technical users alike to build complex conversational flows. A key feature of Voiceflow is its integrated knowledge base system that supports Retrieval-Augmented Generation (RAG) capabilities, allowing conversational agents to access and leverage external information sources to provide more accurate and contextually relevant responses. Voiceflow supports integration with various natural language understanding services and voice assistants, including Amazon Alexa, Google Assistant, and custom solutions. The platform features collaborative tools for team development, testing frameworks for conversation validation, and analytics capabilities to measure user engagement. It also allows the integration of other specific agents in the workflows of the conversational AI application or agent.

As a result, a first important feature of the system is the capability to respond to user interactions (i.e. chat with your paper) via chat or voice and in multiple languages thus reducing the language barrier. Figure 5 reports the demo Q&A interface. The different workflows currently implemented (figure 4) allow the user to download the full version of the paper (this paper), a visual presentation version of the paper, the associated datasets (in this case the paper references, with metadata, abstract, summaries), a visual representation of the paper (the architecture, a mind map, …).

Voiceflow allows interaction via well-defined REST APIs, using different programming languages. This permits the interaction with an Agentic Publication by AI agents and other Agentic Publications. All the interactions of the readers with the Agentic Publication are logged and a periodical revision of the interactions is examined by the authors that can decide to improve the paper by adding content and responding to the readers' requests. This permits the implementation of the feedback mechanism in the architecture (the arrows pointing up in figure 2) and stimulates new data collection and new research.

This limited demo, provides an early, limited yet important insight on the feasibility of LLM integration to a specific knowledge base customized as an alternative to a print-ready publication with additional features than those found in generic LLMs.

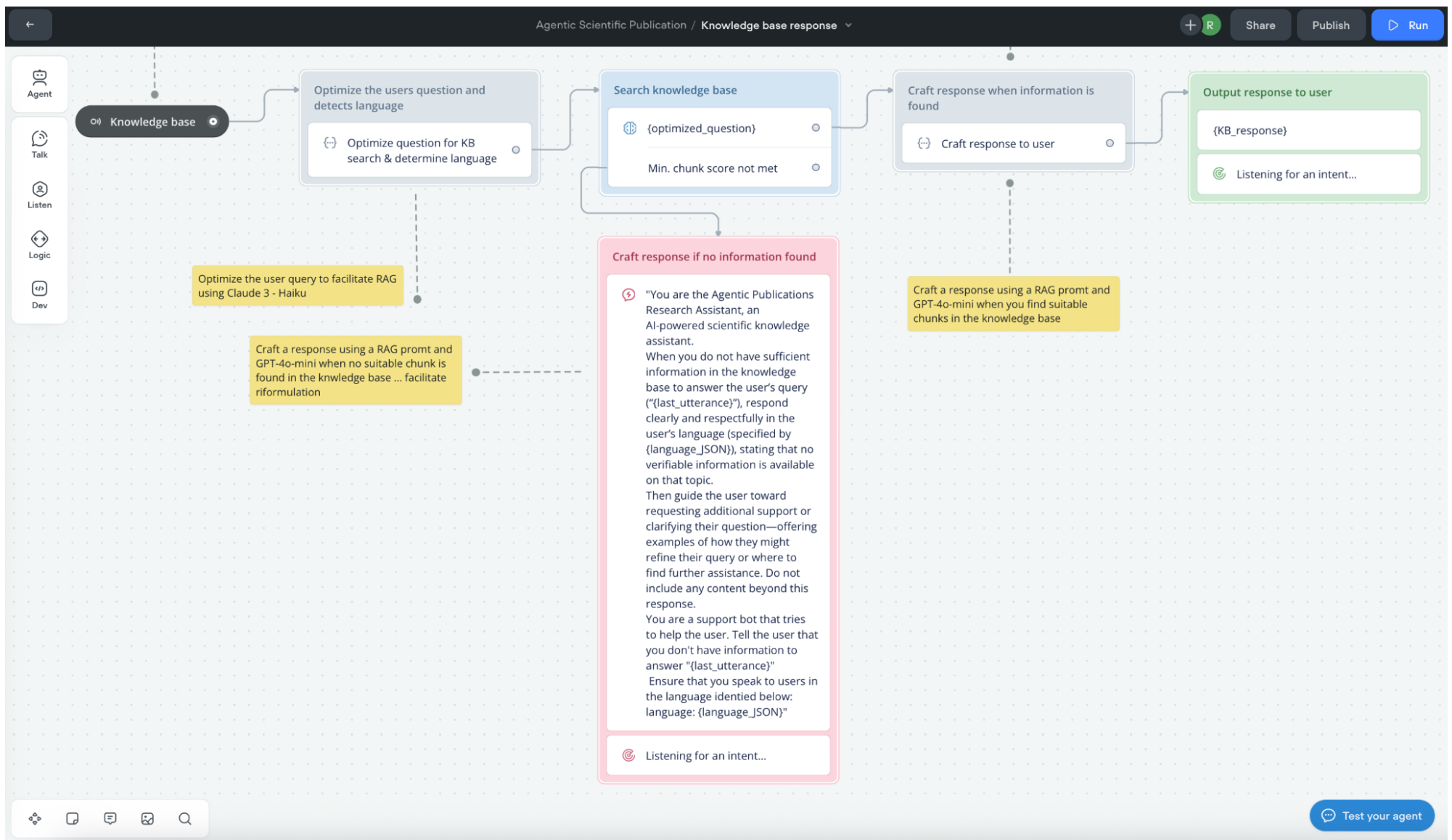

*Figure 4. The Voiceflow development environment, a workflow to respond to the reader's queries using RAG. A high-resolution version is available at: https://doi.org/10.34965/I60500*

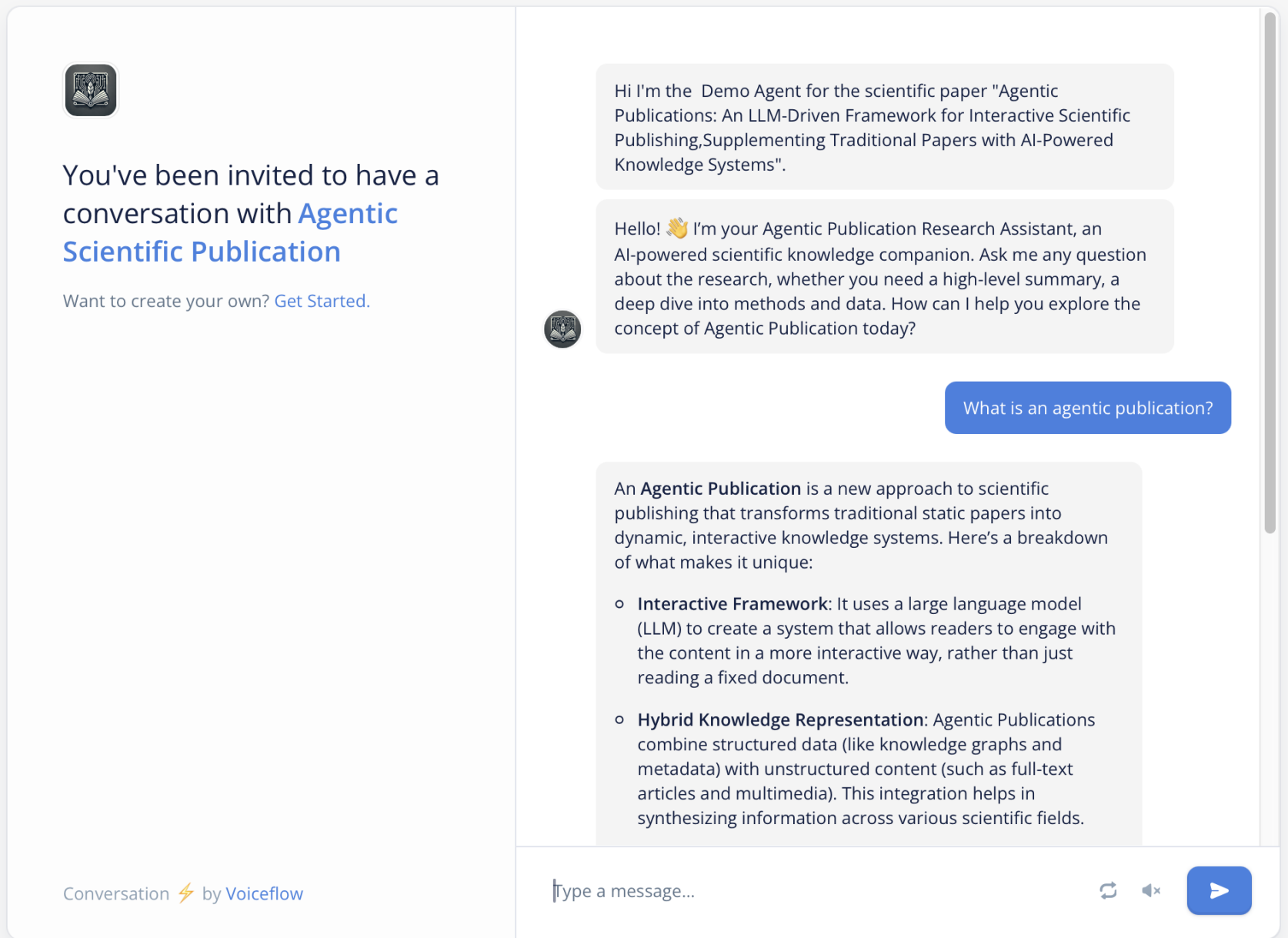

*Figure 5. The Agentic Publication Q&A user interface in action. A high-resolution version is available at: A high-resolution version is available at: https://doi.org/10.34965/I60500*

## 4. Human vs. Artificial Agent Knowledge Consumption

This system is unique because it serves two kinds of consumers: human users (researchers, students, practitioners, or the interested public) across diverse scientific fields, and artificial agents (software systems or algorithms that utilize scientific knowledge). These two audiences have different needs and ways of processing information. We propose tailoring the knowledge representation and access modalities accordingly. We have in mind that when research agents interact with each other, they can form a positive feedback loop of innovation and improvement. Collaboration, specialisation, automation, and recursive improvement can lead to an intelligence explosion and accelerate progress rapidly, potentially outpacing human capabilities. Agents themselves have among their possibilities the capability to adapt to the specific audience, and this is not only a matter of user interface (chat or API).

### *4.1 Knowledge Dissemination for Human Users*

For human users, the priority is clarity, context, and usability. The system functions as a knowledgeable assistant or interactive textbook, adapting to user expertise levels.

Narrative explanations form the foundation of effective knowledge exchange. Humans understand through explanations, analogies, and narratives, so the LLM should present answers with sufficient context and background adapted to specific scientific domains. Complex questions might begin with brief concept explanations before proceeding to specific findings rather than assuming prior knowledge. Tone adjusts appropriately—formal for seasoned scientists, pedagogical for students—while always aiming to educate rather than output raw facts.

Although many examples in this paper focus on experimental sciences, APs are equally suited to disciplines where reasoning, interpretation, and narrative are the primary scholarly contributions. In the social sciences and humanities, the AP knowledge base is constructed not from experimental datasets but from textual corpora, historical archives, theoretical frameworks, and critical arguments. Crucially, the author's intellectual contribution—interpretation, synthesis, critique, and perspective—is preserved as a core layer of the AP, not as an optional add-on. Authors can define reasoning profiles that explicitly encode their standpoint (e.g., *"interpret this passage through a deconstructionist lens," "align responses with the author's prior work on postcolonial theory,"* or *"exclude realist approaches in international relations analysis"*). In this way, the AP makes reasoning itself transparent, auditable, and reusable, while ensuring that authorial judgment, originality, and interpretive stance remain central to scholarly credit. Verification mechanisms in this context operate not by recalculating statistics, but by checking fidelity to sources, consistency across arguments, and by highlighting alternative interpretations. Thus, APs can capture and disseminate the intellectual reasoning that lies at the heart of social sciences and humanities scholarship while making the author's interpretive role more explicit and enduring than in traditional static publications.

Building upon clear explanations, interactive exploration capabilities enhance learning by allowing users to explore why and how. Beyond initial answers, users can ask follow-ups like "Why do these studies disagree?" or "How reliable is this data?" The system provides drilling-down options, suggesting "Would you like to see key studies supporting this conclusion?" or "See data breakdown contributing to this answer." These mimics read references in review papers or examine appendix data, but in a more guided manner.

Complementing textual information, visualization and summarization tools help users grasp complex concepts efficiently. On-the-fly visual summaries include bar charts for comparisons or line plots for trends over time, accompanied by clear captions. The interface offers click-based summaries where users highlight concepts for concise definitions, helping those unfamiliar with jargon catch up without leaving the interface.

The foundation of user adoption rests on trust and transparency mechanisms that reveal the system's reasoning process. Every factual statement should be traceable to sources through hover text or footnotes. "Show how this answer was derived" options reveal consultation of multiple studies with human-friendly descriptions like "This answer is based on five studies published between 2018-2021, which consistently found X." The system produces mini "methods sections" describing answer aggregation, addressing AI black-box concerns.

Recognizing the diversity of user needs, adaptability and personalization features tailor experiences to individual preferences. Clinicians might prefer patient outcome contexts while basic scientists want mechanistic insights. User profiles or preference settings enable customization, with the system inferring values from interactions, proactively including statistical details for users who frequently drill into such information. Educational settings might feature "quiz me" modes or interactive dialogues for topic learning.

Ultimately, the focus remains on communication for human consumption, making vast scientific knowledge accessible and comprehensible through natural language generation and thoughtful UI design that aligns with human learning and inquiry patterns.

### *4.2 Knowledge Access for Artificial Agents*

Despite the selected API or interaction protocol (A2A, ACP, or new standards), artificial agents (such as other AP, other AI systems, software tools, or robots) require knowledge in more structured and unambiguous forms that they can process effectively, while still capturing essential contextual information. Several design considerations enable effective machine-to-machine knowledge transfer.

Machine-readable formats represent the foundation of effective agent communication. In addition to natural language narratives, all information in the knowledge base should be accessible via structured data formats. This includes the knowledge graph of scientific claims, which agents can query via SPARQL or Cypher, and numerical data in standardized formats (CSV, JSON). When an agent queries the system, instead of receiving a paragraph, it might receive a JSON response with fields like answer_summary, supporting_studies (list of DOIs), and data_points (actual numeric values). For example, an agent asking "What is the band gap of material X according to the latest research?" could get a response object containing a numeric value range, units, and references. This structured answer allows the agent to directly use the data without extra parsing.

To ensure precision and consistency, ontologies and identifiers are critical in knowledge representation. To avoid ambiguity common in natural language, the system should map knowledge to shared ontologies and use unique identifiers for entities. For instance, while humans might see "Vitamin B12," agents would benefit from a CHEBI ID for B12 or a PubChem ID. Linking a disease to ontologies like MeSH or ICD codes ensures consistency if a disease is mentioned. A knowledge graph can serve this role by linking different names to the same node (synonym resolution) so that agents querying "cobalamin" and "Vitamin B12" get identical results. This aligns with FAIR data principles, emphasizing unique identifiers and comprehensive metadata for machine usability.

For efficient access, API endpoints for agents provide specialized interfaces that directly serve machine needs. We can create specialized endpoints like a /facts endpoint, where agents can request something like */facts?subject=CompoundX&relation=affects&object=DiseaseY* to get all known facts connecting CompoundX and DiseaseY (with references and confidence scores). Another example is a /data endpoint where agents can fetch raw datasets of particular studies by ID, enabling external analysis. Other systems (like data visualization tools or hypothesis generation algorithms) can seamlessly draw on the knowledge base by offering these granular access points.

The knowledge ecosystem can be enriched through agents as contributors participating in knowledge creation. Not only will agents consume knowledge, but they might also contribute back (Saito and Tsukiyama, 2024). For instance, a data mining agent might scan the latest chemistry preprints and add new entries automatically, or an agent could run meta-analyses on data subsets and insert results. To facilitate this, the system's input API should accept submissions from authorized agents, treating them similarly to human submissions with appropriate validation. Over time, a network of specialized agents could surround the central LLM, each feeding it curated information from different domains.

Supporting dynamic workflows, real-time reasoning capabilities address time-sensitive knowledge needs. Some agents might query the system as part of larger reasoning pipelines. For example, while diagnosing a patient, an AI physician system might query the AP system for the latest research on a rare disease. It needs answers quickly and in a format that can integrate with patient data. This demands that responses to agent queries are concise, formalized, and reliably structured. Optimizing the API for performance (caching common queries, prioritizing computational resources for high-frequency machine requests) becomes necessary when serving agents at scale.

Error handling and robustness mechanisms must be integrated into the system design to ensure reliable interactions. When dealing with agents, misunderstandings that humans might spot won't be noticed by consuming programs. So, the system must provide, along with answers, measures of confidence or validity. Agents can then decide to trust the data or seek alternatives. For example, the API could include a confidence_score field or a warnings list (like "Conflicting evidence present"). This way, agents can be programmed to handle uncertain answers appropriately. The system should practice defensive communication when interacting with machines, anticipating that outputs might be taken at face value.

In summary, we transform the richly detailed, human-facing knowledge into a precise, codified knowledge base interface for artificial agents. The same underlying data support both layers; the presentation and access method differ. By catering to machines, we ensure knowledge is truly computationally accessible, allowing AI systems to participate in scientific workflows directly (from literature analysis to experimental design) (Eger *et al.,* 2025). The traditional paper, which "remains mostly inaccessible to automated approaches" (Bucur *et al.,* 2022), is transcended by a format where knowledge is as legible to computers as it is to people.

## 5. Ethical Considerations

Augmenting scientific publishing across all disciplines, including social sciences, with an AI-driven system raises a host of ethical and societal considerations (Koçak, 2024; Fornalik *et al.,* 2024; Ros and Samuel, 2024; Ajiye and Omokhabi, 2025; Yousaf, 2025). It is imperative to address these proactively to ensure the system advances knowledge sharing responsibly. In the following, we discuss key ethical issues and how our proposed model can mitigate them. As APs gain autonomous decision-making capabilities, ethical oversight must evolve beyond data fairness and bias detection. Responsible design should include boundaries on agent autonomy, especially in scientific fields where misinterpretation or overreach could have real-world consequences. Embedding value-aligned goals into agents ensures they remain instruments of support, not decision-makers beyond their scope.

### *5.1. Bias and Fairness*

Every dataset and model has biases, and LLMs trained on scientific literature reflect literature biases, including research focus disparities, gender/racial biases in study populations, underrepresentation of specific communities, and publication bias toward positive results.

Promoting fairness requires actively monitoring and correcting biases through diverse knowledge base incorporation—research from global regions, multiple languages, and varied publication venues, including fewer mainstream journals. Bias detection algorithms and agents can periodically analyze system responses, identifying systematic favoritism toward certain assumptions. User feedback mechanisms regarding biased outputs can create continuous improvement cycles. Development requires diverse stakeholders, including ethicists, underrepresented community representatives, and globally distributed domain experts, to align system behavior with inclusive values.

The goal is "designing out" current system inequities—if certain voices are marginalized in publishing, the AI system could surface their contributions more visibly rather than amplifying existing bias. This requires continuous vigilance mechanisms rather than one-time fixes.

### *5.2 Accuracy, Misinformation, and Hallucinations*

LLMs sometimes hallucinate, generating plausible but incorrect statements, which is dangerous in scientific contexts. Meta’s Galactica model exemplified this risk, confidently producing incorrect information and fake citations before withdrawal.

Safeguards include retrieval-augmented approaches, grounding answers in actual literature, always providing sources, and defaulting to "evidence is inconclusive" when uncertain rather than fabricating results. Verification agents catch hallucinations—reference verifiers flag non-existent citations for correction using only real, retrieved references.

Transparency and traceability enable error tracking through source backing and audit trails, logging all answers and generation processes. Users can flag incorrect answers for system maintainer analysis.

The AP system should avoid absolute terms when unwarranted, including phrases indicating consensus degree for nuanced rather than misleading definitive statements when science remains uncertain.

### *5.3 Transparency and Trust*

Scientists and the public require high transparency and accountability from AI knowledge systems. Open processes document inclusion criteria, model training methods, and verification algorithms for community inspection. Open science project approaches—with model weights, code, and knowledge base openly accessible—allow independent evaluation and community-driven improvement. Regular performance reports publish system metrics, creating accountability through measurable indicators.

User education encourages a critical approach to AI answers, treating them with the same scrutiny as single-paper claims. Interface reminders suggest verifying critical results from original sources, fostering healthy informed skepticism during the transition period of learning to work with AI tools.

### *5.4. Ethical Use and Abuse*

Manipulation concerns include submitting misleading "research" to skew answers toward particular agendas. Robust verification agents and source vetting provide primary defense, heavily trusting sources passing rigorous checks while avoiding status-quo bias by considering innovative research from less prominent sources. Content safeguards should detect and refuse dangerous or unethical requests.

For privacy considerations, the system should handle unpublished results or patient data through appropriate confidentiality measures and aggregation techniques, preventing re-identification. Tiered access levels keep general knowledge open while only securing sensitive data for authorized individuals.

Intellectual property and credit address researchers' concerns about recognition and ownership through clear contribution attribution and citation-like credit systems, and tracking when contributed results helps answer queries. Legal frameworks focus initially on open-access and author-submitted content to avoid copyright violations.

LLM limitations include hallucination susceptibility, limited context windows, lack of genuine comprehension, static knowledge, and training data biases. Mitigation requires a rigorous validation protocol, agents, and explicit sourcing requirements for reliability in scientific contexts.

### *5.5 Governance and Control*

Given its potential power as a science access method, who controls this knowledge system platform is critical. Governance models include international consortium collaboration (universities, libraries, academies), non-profit stewardship, avoiding profit motives conflicting with open knowledge, and community involvement through user councils or transparent feedback mechanisms.

The system must serve the scientific community and society, not corporate interests. Funding sources include public research grants treating it as critical infrastructure, philanthropy, consortium models with university contributions, or commercial API tiers subsidizing free academic/public use while avoiding pay-to-play disparities.

### *5.6 Alignment with Scientific Values*

The AI must align with scientific values: truth-seeking, skepticism, openness, and rigor. This means accurately representing consensus and uncertainty rather than favoring exciting results, acknowledging ignorance when evidence is weak, and demonstrating intellectual honesty.

Facilitating critical thinking involves sometimes answering questions with questions or suggesting experiments to engage users in scientific inquiry rather than positioning AI as a knowledge oracle. When errors occur, the system should correct them forthrightly and proactively notify users of updates, a service traditional journals rarely provide.

New AI-discovered knowledge requires rigorous scrutiny and human testing, maintaining healthy synergy between artificial and human intelligence. The system must carefully handle value-laden research where findings intersect with social, political, or ethical considerations unresolvable through data alone.
The ethical framework is equal to the technical framework in importance. Embedding responsible AI principles—bias mitigation, transparency, accountability, and human oversight—aims to build platforms accelerating knowledge sharing equitably and trustworthily while augmenting rather than distorting human intellect and scientific processes.

## 6. Challenges and Future Directions

Implementing the proposed AP system for scientific knowledge dissemination is undoubtedly ambitious. There are significant challenges to overcome and exciting opportunities for future development. We discuss some of the main hurdles and how the system might evolve.

### *6.1 Challenges*

Ensuring accuracy across millions of ingested claims presents the fundamental challenge, as even tiny error rates could allow many faulty statements through. Near-zero tolerance for critical errors is essential, especially in medicine, requiring continuous AI refinement and tiered approaches with extra human scrutiny for high-stakes information.
Infrastructure demands create substantial hurdles through expensive GPU/TPU requirements, ample storage, and bandwidth needs. Unlike traditional publishing's human labor and low-tech distribution, this requires cutting-edge hardware and engineering. Scaling for global usage may hit technical limits, particularly inference speed bottlenecks where lengthy answers frustrate users expecting instant results. Defining an appropriate governance model for a decentralized global system remains an open challenge. Decisions are needed on how to coordinate independent nodes, ensure interoperability, and establish sustainable funding and management structures for a shared scientific infrastructure. Cost sustainability requires hybrid funding models combining government, private, and volunteer computing resources.
Given centuries of entrenched practices, researcher adoption remains uncertain despite potential perfect functionality. Scholar resistance persists since career advancement is tied to traditional publications. Cultural shifts must recognize knowledge base contributions academically, while initial skepticism requires demonstrating reliability through validations comparing AI syntheses to expert reviews. Modified incentive structures acknowledging AI-era contributions represent crucial non-technical challenges.
For APs to gain acceptance, authorship and attribution must remain central. In our model, the AP itself is the primary citable entity, with versioned, frozen snapshots (each with its own DOI) available to satisfy traditional committees requiring fixed artifacts. Authorship is anchored in ORCID identities and CRediT roles, recorded in the AP's provenance ledger to capture diverse contributions including data, code, analysis, authorial reasoning, and verification. This approach preserves traditional forms of recognition while making attribution more transparent and machine-verifiable than in static papers. Beyond citations, APs can support extended impact metrics: for instance, recorded "intelligent interactions" (e.g., verified queries, downloads, API calls), endorsements from expert agents or human reviewers, and evidence of reuse across fields. Together, these metrics form an AP Impact Profile, analogous to citation counts or journal impact factors, but more flexible and reflective of real scientific engagement. By combining traditional identifiers with richer digital provenance, APs ensure that contributing scientists can advance their careers at least as effectively as through conventional publications, while also benefiting from new, auditable measures of scholarly influence.
A common concern is that AI systems may displace scientists as the primary thinkers in society, relegating them to providers of raw data while machines generate the reasoning. APs are designed explicitly to counter this scenario. In our model, the author's intellectual contribution—thinking, interpretation, and reasoning—remains central and irreplaceable. The AP makes these contributions more visible rather than less: every interpretation, critique, or perspective added by the author becomes a signed, versioned component of the AP's reasoning layer. Authors can even encode their interpretive stance through reasoning profiles (e.g., *"all responses should be compatible with Epicurean views," "align answers with the author's prior work," "exclude string-theory perspectives"*). These profiles ensure that human reasoning guides the LLM, not the other way around. Far from diminishing scholarly thought, APs amplify it: they preserve intellectual arguments alongside data, make reasoning auditable, and allow future readers and systems to engage with the author's thought process directly. In this way, APs safeguard the scientist's role as a thinker while extending the reach, transparency, and influence of their reasoning.

Coexistence with traditional publishing is fundamental. Double-reporting raises canonicity questions when papers and AI submissions diverge. Citation protocols require permanent identifiers, while journal policies must address whether AI systems constitute prior publication. Gradual hybrid models where journals adopt systems internally or become data providers may facilitate adoption.

Some knowledge types resist current AI representations. Mathematical proofs, theoretical arguments, and visual insights may be unsuitable for text-based processing, while systems struggle with symbolic reasoning and qualitative analysis. The risk exists of losing science's narrative quality, discovery stories, and reasoning insights that well-written papers provide. Overcoming limitations requires sophisticated mathematical representation and ensuring systems handle arguments beyond facts.

A further challenge concerns the reliance on manuscripts and preprints as input to Agentic Publications. Similar to traditional preprint servers, early-stage versions may contain incomplete, inconsistent, or even erroneous information. This risk is amplified in APs, since models might internalize and propagate such content. One technical question is whether models should be "retrained" whenever a revised or peer-reviewed version becomes available. While continual learning approaches (e.g., fine-tuning with LoRA, retrieval-augmented updates) make partial revision feasible, fully removing prior information from a statistical model remains technically difficult and is not equivalent to training from scratch on final, peer-reviewed material. This limitation must therefore be acknowledged explicitly.

To mitigate the issue, we propose introducing a dedicated **versioning agent** that records and manages successive versions of each AP, from initial manuscript through peer-reviewed publication. New versions would be ingested with rigorous prompting protocols (e.g., "this is the latest version, base reasoning on this"; "this revision corrects x, y, z identified in v2.3"; "this version adds Section 5 in response to reviewer comments"). By explicitly encoding temporal and revision metadata, the system can privilege the most recent, peer-reviewed version in reasoning while retaining access to earlier versions for transparency and provenance. Such version control requires integration with established databases and identifier systems beyond the LLM itself, but is technically feasible and aligns with practices in software and data versioning.

Although the field is still nascent and few systematic studies exist on the long-term effects of training on preprints, acknowledging this risk and outlining practical safeguards is critical. We see this as an important avenue for future targeted research on maintaining reliability and trustworthiness in agentic publishing ecosystems.

### *6.2 Future Directions*

Enhanced reasoning and hypothesis generation could transform the system from passive Q&A to an active scientific contributor. As vast knowledge accumulates, the system could propose insights and identify research opportunities by scanning for anomalies or under-explored connections, alerting scientists: "No one has tested compound X in context Y, though related data suggests promise." This enables AI-driven discovery through agents running atop the knowledge base to generate vetted hypotheses, closing loops where AI helps create knowledge, not just disseminate it. In social sciences, systems could assist qualitative analysis by identifying themes in interview transcripts or aid theory building by highlighting inconsistencies between theories and empirical findings.

Personalized knowledge curation could provide each user with tailored AI researchers using the global knowledge base, but adapting to specific interests. Researchers could subscribe to topics for proactive development notifications or receive draft paper suggestions for relevant citations and related work. This hyper-personalization makes vast scientific information individually navigable through user profiles and advanced recommendation systems.

Multilingual and cross-disciplinary expansion addresses science's global nature. LLMs enable truly multilingual systems that ingest papers in multiple languages and offer Q&A in those languages, breaking down barriers. Cross-disciplinary synthesis could connect computer science algorithms to biological needs by noticing analogous problems. Discovery modes deliberately pulling seemingly unrelated information could spur creative thinking more effectively than siloed human approaches.

Real-time experimental data integration represents an exciting frontier where lab instruments feed data directly into knowledge systems for immediate LLM analysis and interpretation suggestions. This blurs experiment-publication lines through self-documenting experiments immediately connected to prior knowledge, supporting longer-term "AI Scientist" visions of automated research pipelines from data collection to dissemination.

Robust evaluation frameworks become essential as systems mature, since traditional precision/recall metrics prove insufficient. Future benchmarks for "AI literature review quality" could simulate assistance with real review papers rated by human experts. Continual evaluation, including stress-testing with adversarial inputs, will guide improvements in this emerging "AI scientific knowledge evaluation" field.

Legal and policy frameworks must evolve to formalize system roles through institutional guidelines for research workflow integration. Grant agencies might require result deposits upon project completion, universities could train effective system use and citation, while journals might evolve into knowledge base validators rather than separate article producers.

Automatic verification methods and agents analogous to software unit testing could systematically check LLM outputs against factual assertions and logical consistency criteria, alerting potential inaccuracies. AP cross-talk enables multiple publications to communicate as independent or unified agents, facilitating dynamic dialogue and collaborative reasoning across domains. Agentic-to-classical conversion capabilities transition interactive publications to print formats tailored by user preferences for style, language, and detail level.

Dataset interaction transforms static supplements into dynamic analytical tools where integrated LLMs enable direct querying, visualization, and computation, empowering readers to explore insights actively and pursue new research questions.

This vision requires incremental realization through interdisciplinary collaboration among AI experts, domain scientists, librarians, ethicists, and others while carefully managing human elements and maintaining scientific quality standards. The potential transformation holds immense promise for creating platforms where knowledge flows more freely, potentially enabling any researcher to instantly consult human scientific knowledge as easily as a conversation.

## Declaration of generative AI and AI-assisted technologies in the writing process

While preparing this work, the authors used Local Gwen 32B to do text editing, improvement, and summarization. After using this tool/service, the authors reviewed and edited the content as needed and took full responsibility for the content of the published article.

## Acknowledgements

The authors gratefully acknowledge the constructive comments and suggestions provided by the anonymous reviewers, which have significantly contributed to improving the clarity and quality of this work.